\documentclass{article}
\pdfoutput=1
\usepackage[sc,osf]{mathpazo}
\usepackage[height=8.5in,width=6in,letterpaper]{geometry}
\usepackage[tracking]{microtype}
\usepackage{amsfonts}
\usepackage{amsmath,amssymb,amscd,epsfig,amsfonts,rotating}
\usepackage{multirow}
\usepackage{booktabs}
\usepackage{xcolor}
\usepackage{url}
\usepackage{authblk}
\usepackage{epsfig,epstopdf,graphicx,subfigure}
\usepackage{enumitem,balance}
\usepackage{wrapfig}
\usepackage{mathrsfs, euscript}
\usepackage{algorithm}
\usepackage{algorithmic}
\usepackage{ifpdf}
\usepackage{makecell}
\usepackage[colorlinks=true,citecolor=blue,breaklinks]{hyperref}
\usepackage{appendix}

\title{TripletGAN: Training Generative Model with Triplet Loss}

\author[1, 4]{\small Gongze Cao}
\author[2]{\small Yezhou Yang}
\author[1]{\small Jie Lei}
\author[3]{\small Cheng Jin}
\author[5]{\small Yang Liu}
\author[1]{\small Mingli Song}    

\affil[1]{\footnotesize Zhejiang Provincial Key Laboratory of Service Robot, Zhejiang University, China}
\affil[2]{\footnotesize School of Computing, Informatics, and Decision Systems Engineering, Arizona State University}
\affil[3]{\footnotesize School of Computer Science, Fudan University, Shanghai, China}
\affil[4]{\footnotesize School of Mathematics, Zhejiang University, Shanghai, China}
\affil[5]{\footnotesize Alibaba Group, Hang Zhou, China}

\date{}
\begin{document}

    \maketitle
    \begin{abstract}
        As an effective way of metric learning, triplet loss has been widely used in many deep learning tasks, including face recognition and person-ReID, leading to many states of the arts. The main innovation of triplet loss is using feature map to replace softmax in the classification task. Inspired by this concept, we propose here a new adversarial modeling method by substituting the classification loss of discriminator to triplet loss. Theoretical proof based on IPM (Integral probability metric) demonstrates that such setting will help generator converge to the given distribution theoretically under some conditions. Moreover, since triplet loss requires the generator to maximize distance within a class, we justify tripletGAN is also helpful to prevent mode collapse through both theory and experiment.
    \end{abstract}

    \section{Introduction}
    The generative model has been studied thoroughly throughout these years, \cite{salakhutdinov2009learning,bengio2013generalized}. Among them Generative Adversarial Networks(GAN) \cite{goodfellow2014generative} has proved its superiority in many tasks, such as image generation, style transfer, 3D object modeling, and image super-resolution \cite{radford2015unsupervised,berthelot2017began,zhu2017unpaired,wu2016learning}. GAN adopts a different training process with many previous generative models, it proposes to learn a parametrized distribution(denoted by $G(z)$) through an auxiliary classifier $D$ called discriminator which try to discriminate between true and fake samples. Its training process hence consists two steps, first get the probability of current sample using softmax function, then update $D$ with cross-entropy loss with respect to the label of sample, secondly, update $G$ by maximizing the probability of its samples to be true(or some variants by applying a monotonic increasing function on). It's proved by \cite{goodfellow2014generative} that this algorithm results in minimizing the Jensen-Shannon divergence between the data distribution and the generated distribution under some condition. Many following works generalize it to a larger class of divergence such as f-divergence \cite{nowozin2016f}.

    Though GAN enjoys successful applications in many fields, it is well-known that its training suffers from many issues, including the instability between generator and discriminator, and the extremely subtle sensitivity to network architecture and hyperparameters. \cite{arjovsky2017towards} showed with theoretical proof that most of the training problems of GAN are due to the fact that the support of both target distribution and generated distribution are often of low dimension regarding to the base space, hence misaligned at most of the time, causing discriminator to collapse to a function that hardly provides gradients to generator. In this case, the usually minimized divergence (such as KL, JS) will be raised to infinity. To remedy this issue, \cite{arjovsky2017wasserstein} propose to minimize Wasserstein distance between data and generated distribution, which can also be interpreted in IPM (Integral Probability Metric) form.

    Another particular common problem of original GAN is mode dropping, namely, it refers to the phenomenon that during training generator tends to emit high probability samples from a limited number of modes. A lot of works have been done to solve this problem, such as minibatch discrimination \cite{salimans2016improved}, unrolled GAN \cite{metz2016unrolled}, mode regularized GAN \cite{che2016mode}. Instead of introducing a regularizer on the generator, we integrate a tendency term for the generator to produce diverse samples by utilizing triplet loss. Triplet loss is widely used in face recognition \cite{schroff2015facenet} and metric learning \cite{hoffer2015deep}, in place of the classification based method, which requires producing a large dimension output to perform softmax on. Triplet loss method only requires to map to a fixed low-dimension space and then minimizes the distance between embeddings of the same class, meantime maximizes that of the different class. It is easy to convert vanilla GAN to our proposed tripletGAN by simply substituting softmax to an embedding map, and classification loss to triplet loss, by analogy. We show in following sections that our proposed tripletGAN guarantee theoretically that the generated distribution is able to converge to real distribution and help to prevent the mode collapse problem by updating generator to maximize the embedding distance between fake samples. Furthermore, we conduct several experiments on mode recovering and image generation in contrast to vanilla GAN, showing our method leads to better mode coverage.
    \section{Related work}
    \subsection{Minibatch Discrimination}
    Minibatch Discrimination was proposed in \cite{salimans2016improved} to avoid the common problem that the generator always emits the same point and discriminator posit no punish on this situation. It introduced a regularizer that explicitly maximizes the distance of features in a minibatch, so when fake samples are in the same mode, the regularizer would provide gradients for samples in a minibatch to differ with each other. \cite{zhao2016energy} provide a similar method called repelling regularizer that punishes the cosine similarity of the feature in the same batch, forcing the features to be orthogonalized pairwise.

    Our method is similar with these work in that the latter part of the loss for the generator to minimize is exactly the cosine similarity between the embedding vectors of fake samples, but at the same time, it also serves as part of modeling, not just a regularizer. So critic\footnote{In original GAN the auxiliary classifier is often called discriminator, while in Wasserstein GAN and some IPM based GAN called the critic. Since both of them are designed to provide adversaries, we do not try to distinguish this two concepts in this paper.} also learns to map fake samples to the same point, which allows a more meaningful measure of closeness than simply utilizing features trained in other objectives.
    \subsection{Maximum Mean Discrepancy}
    In general, MMD (Maximum Mean Discrepancy) tell the difference between two distribution by comparing the mean embeddings of samples from two distributions. The MMD between two distributions $P_A$ and $P_B$ with an embedding function $\varphi:\mathcal{X}\to\mathcal{H}$ is:
    \[MMD(P_A, P_B)=\lVert\mathop{\mathbb{E}}_{x\sim P_A}[\varphi(x)]-\mathop{\mathbb{E}}_{y\sim P_B}[\varphi(y)]\rVert_{\mathcal{H}}\ ,\]
    where $\mathcal{X}$ is the space on which the distribution is defined, $\mathcal{H}$ is the feature space, often taken as a reproducing kernel Hilbert space (RKHS), to get a closed form in terms of its corresponding kernel $k$. Note that MMD can be seen as a particular class of IPM, and it has been used in distribution modeling in many previous works \cite{dziugaite2015training,li2015generative,sutherland2016generative}.

    Similiar to minimizing MMD, McGAN \cite{mroueh2017mcgan} also proposed to learn the target distribution by matching the mean encodings of two minibatch sampled from the different distribution. But rather than utilizing kernel trich to obtain a closed form estimation of distance, it uses adversarial learning to learn the feature mapping function $\varphi$. Moreover, it present that it is also useful to matching higher order statistics of encodings.

    Though both McGAN and ours map samples to a high dimension encodings, The mean feature matching GAN proposed in McGAN first takes the mean encodings across a minibatch, and then operate on them, while ours treat the encodings as triplets. We first operate on triplets and then take the mean of them. Through a toy setting we are able to probe it clearly: suppose now we have an encoding function $\varphi$ for $\mathcal{P_{A}}$ and $\mathcal{P_{B}}$, and $\varphi$ transform $\mathcal{P_{A}}$ and $\mathcal{P_{B}}$ to two gaussian distribution $\mathcal{N}(0, \sigma_1^2)$, $\mathcal{N}(0, \sigma_2^2)$, $\sigma_2\ge \sigma_1$, for ease of denoting, we refer $X$ and $Y$ as the random variables corresponding to $\mathcal{P_{A}}$ and $\mathcal{P_{B}}$ separately, then the distance by mean encoding matching would yield:
    \[
     \small\begin{split}
        d_{me} &= \lVert\mathop{\mathbb{E}}_{x\sim\varphi(X)}x-\mathop{\mathbb{E}}_{y\sim\varphi(Y)}y\rVert\\ &=\lVert\mathop{\mathbb{E}}_{x\sim\mathcal{N}(0, \sigma_2^2)}x-\mathop{\mathbb{E}}_{y\sim\mathcal{N}(0, \sigma_1^2)}y\rVert\\&=0
     \end{split}
    \]
    While our proposed tripletGAN would minimize:
    \[
     \small\begin{split}
        d_{tr} &= \left|\mathop{\mathbb{E}}\limits_{\substack{y\sim\varphi(Y),\\ x_1\sim\varphi(X)}}\lVert y-x_1\rVert-\mathop{\mathbb{E}}\limits_{x_1, ~x_2\sim\varphi(X)}\lVert x_1-x_2\rVert\right|\\ &=\left|\mathop{\mathbb{E}}\limits_{\substack{y\sim\mathcal{N}(0, \sigma_1^2),\\ x_1\sim\mathcal{N}(0, \sigma_2^2)}}\lVert y-x_1\rVert-\mathop{\mathbb{E}}\limits_{x_1, ~x_2\sim\mathcal{N}(0, \sigma_2^2)}\lVert x_1-x_2\rVert\right|\\&=\left|\sqrt{\dfrac{2}{\pi}}(\sqrt{\sigma_1^2+\sigma_2^2}-\sqrt{2}\sigma_2)\right|
     \end{split}
    \]
    So our critic would provide gradients for the generator to match the varience of encodings, like what is suggested in McGAN, matching higher order statistics. While simply matching the mean of encodings gives no meaningful information.

    \section{Preliminaries}
    \subsection{Integral Probability Metrics}
    IPM (Integral Probability Metrics) is a metric defined on probability space \cite{muller1997integral}. A metric is a bilinear function satisfying several properties, including positive-definite, symmetry and triangle inequality. Let $(\mathcal{X}, \mathscr{S})$ be a probablity space, $\mathcal{F}$ be a function set that is measurable and defined on $X$, then for two arbitrary distribution $P$ and $Q$ in $(\mathcal{X}, \mathscr{S})$, the IPM distance between them is:
    \[
        d_{\mathcal{F}}(P, Q):=\sup\limits_{f\in \mathcal{F}}\left|\int f dP - \int f dQ\right|
    \]
    Note when $\mathcal{F}$ is symmetric, i.e. both $f$ and $-f$ belongs to $\mathcal{F}$, the absolute value could be eliminated. Since most of the time this condition stands, $d_{\mathcal{F}}$ could be written as:
    \[
        d_{\mathcal{F}}(P, Q):=\sup\limits_{f\in \mathcal{F}} \left\{\int f dP - \int f dQ\right\}
    \]
    Actually, the integral form already guaranteed itself a pseudo-metric over $(\mathcal{X}, \mathscr{S})$. To use it as a metric, we only need to choose $\mathcal{F}$ to be large enough to make it positive-definite, that is, $d_{\mathcal{F}}(P, Q)$ inplies $P=Q$. Many such $\mathcal{F}$ were proposed with various interesting properties \cite{sriperumbudur2009integral}. And a lot of works have been done to integrate those metrics into generative models \cite{arjovsky2017wasserstein,dziugaite2015training,li2015generative}. Such as choosing $\mathcal{F}$ to be all lipshitz-1 measurable functions \cite{arjovsky2017wasserstein}, which helps to mitigate the notorious problems of gradient missing and imbalanced training of GAN.
    \subsection{Triplet Loss}
    \label{tripletloss}
    Triplet loss was introduced in \cite{weinberger2009distance}, and improved in many works like \cite{schroff2015facenet}. It was used to find an embedding function which maps data with same label to be close in embedding space, and data of different classes to be far from each other. For a triplet dataset $\mathbb{T}=\{t_i\}$ composed of triplets like $t_i=(x_i^a,x_i^p,x_i^n)$, where $x_i^a$ and $x_i^p$ are of the same class, $x_i^n$ is of different class, triplet loss can be written as:
    \[
        \sum_{i}^N\lVert f(x_i^a)-f(x_i^p)\rVert-\lVert f(x_i^a)-f(x_i^n)\rVert
    \]
    Without specification, all $\lVert\cdot\rVert$ in this paper refers to L2 norm.
    Usually, the per-triplet loss below some threshold $a$ was not taken into account. In other words, we want distance between embeddings of distinct classes to be greater than the in-class distance by at least $a$, but not too great to affect other classes. In this case, triplet loss can be represented as:
    \[
        \sum_{i}^N\left[\lVert f(x_i^a)-f(x_i^p)\rVert-\lVert f(x_i^a)-f(x_i^n)\rVert+a\right]_+\ ,
    \]
    where $a$ is the threshold, $\left[\cdot\right]_+$ refers to $\max(\cdot, 0)$.

    There has been work that tries to integrate triplet loss with adversarial modeling \cite{zieba2017training}, but its main focus is to improve triplet network with the leverage of discriminator, while our concentration is on the training a generative model.
    \begin{figure*}[t]
        \vspace{-4pt}
        \centering
        \includegraphics[width=0.90\textwidth]{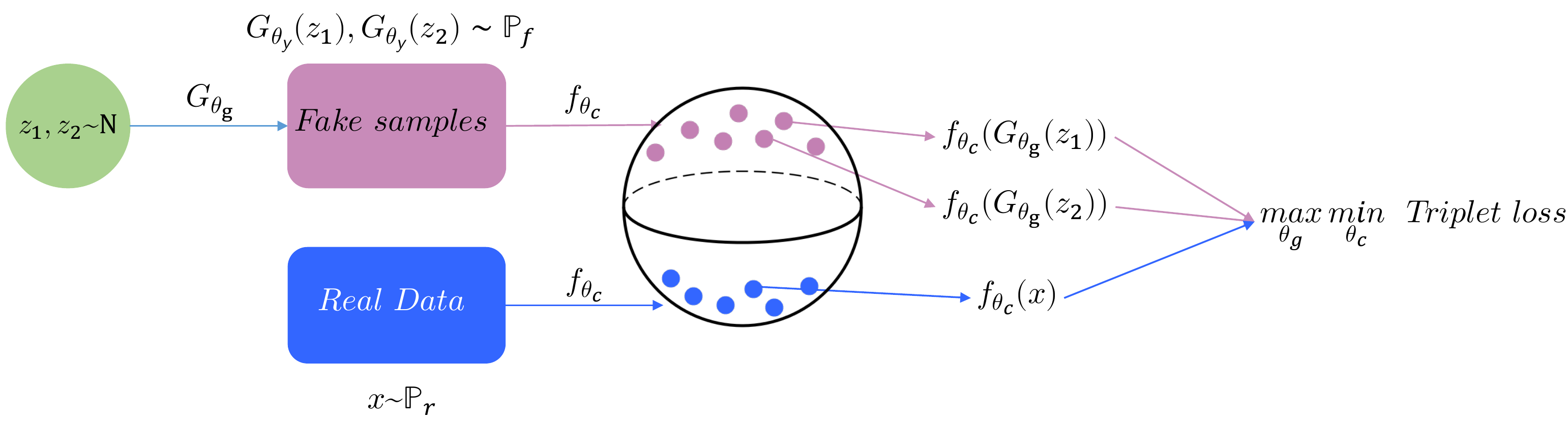}
        \vspace{-8pt}
        \caption{The architecture of triplet GAN. $\mathbb{N}$ stands for the noise distribution. $G(z_1)$ and $G(z_2)$ obey to $\mathbb{P}_f$, $x$ obeys to $\mathbb{P}_r$. $f$ and $G$ are parametrized to $\theta_g$ and $\theta_c$ respectly. $f$ maps both $G(z_i)$ and $x$ to a high dimension sphere, then $G$ and $f$ play a min-max game with objective to be the triplet loss calculated.}
        \label{fig:tripletGAN-illust}
        \vspace{-10pt}
    \end{figure*}
    \section{Adversarial modeling using triplet loss}
    \subsection{definition}
    Say now we have a target distribution $\mathbb{P}_{r}$ , and a generated distribution as $\mathbb{P}_{f}$ on $(\mathcal{X}, \mathscr{S})$ . Different with the what is stated above about triplet loss, where often thousands of classes are of interest, we only have two classes under consideration here, that is true and fake \footnote{Throughout this paper, we will use true samples and fake samples to denote samples from data distribution $\mathbb{P}_{r}$ and generated distribution $\mathbb{P}_{f}$ respectly.}. We choose fake classes to be anchor and positive class, and true class to be negative class. We follow the naming tradition of GAN to denote $f$ as critic, and $G$ as generator which meant to generate samples that resembles true samples from noise $z$, that is, a transform function map from $z$ to $\mathbb{P}_{f}$, where $z\sim \mathbb{N}$, $\mathbb{N}$ is the noise distribution. Denote $f$'s parameter as $\Theta_c$, $G$'s parameter as $\Theta_g$, then we are able to formulated tripletGAN as a min-max problem that much resembles many GAN variants:
    \[
        \min_{\Theta_g}\max_{\Theta_c}\mathscr{L}_t\ ,
    \]
    where
    \[
        \small\begin{split}
        \mathscr{L}_t&=\mathop{\mathbb{E}}_{\substack{y\sim\mathbb{P}_{r}\\ x_1,~x_2\sim\mathbb{P}_{f}}}(\lVert f(y)-f(x_1)\rVert-\lVert f(x_1)-f(x_2)\rVert)\\
        &=\mathop{\mathbb{E}}\limits_{\substack{y\sim\mathbb{P}_{r},\\ x_1\sim\mathbb{P}_{f}}}\lVert f(y)-f(x_1)\rVert-\mathop{\mathbb{E}}\limits_{x_1, ~x_2\sim\mathbb{P}_{f}}\lVert f(x_1)-f(x_2)\rVert\\
        &=\mathop{\mathbb{E}}\limits_{\substack{y\sim\mathbb{P}_{r},\\ z_1\sim\mathbb{N}}}\lVert f(y)-f(G(z_1))\rVert-\mathop{\mathbb{E}}\limits_{z_1, ~z_2\sim\mathbb{N}}\lVert f(G(z_1))-f(G(z_2))\rVert\\
        \end{split}
    \]

    With a slight abuse of notation, we denote $G(z)$ as the random variable associated to $\mathbb{P}_f$.

    It's ready to see this objective agree with the form of IPM. In fact, denote $S^n$ as the n-sphere, defined by $S^n=\left\{x\in \mathbb{R}^{n+1}:\lVert x\rVert=1\right\}$, then take $\mathcal{F}=\{f\mid f(x): \mathbb{R}^m\mapsto S^n , f\ is\ measurable\}$, $\mathcal{T}=\{\lVert f(x)-f(y)\rVert \mid f\in \mathcal{F} \}$, $\mathcal{X}=\mathbb{R}^m\times\mathbb{R}^m$, the IPM on $(\mathcal{X}, \mathscr{S})$ between two arbitrary distribution $\mathbb{P}$ and $\mathbb{Q}$ with respect to $\mathcal{T}$ is:\\
    \[
        \small\begin{split}
            d_{\mathcal{T}}(\mathbb{P}, \mathbb{Q})&=\sup\limits_{g\in \mathcal{T}}\left\{\mathop{\mathbb{E}}_{(x,~y)\sim \mathbb{P}}g(x, y)-\mathop{\mathbb{E}}\limits_{(x,~y)\sim\mathbb{Q}}g(x,y)\right\}\\
            &=\sup\limits_{f\in \mathcal{F}}\mathop{\mathbb{E}}_{(x,~y)\sim \mathbb{P}}\left\{\lVert f(x)-f(y)\rVert-\mathop{\mathbb{E}}\limits_{(x,~y)\sim\mathbb{Q}}\lVert f(x)-f(y)\rVert\right\}
        \end{split}
    \]
    Let $\mathcal{P}$ equals to the independent joint distribution of $\mathbb{P}_{r}$ and $\mathbb{P}_{f}$, $\mathcal{Q}$ equals to the independent joint distribution of $\mathbb{P}_{f}$ and $\mathbb{P}_{f}$, so it's obvious that the the objective of tripletGAN is minimizing the IPM distance between the independent joint distribution of $\mathbb{P}_{r}$ and $\mathbb{P}_{f}$ and the independent joint distribution of $\mathbb{P}_{f}$ and $\mathbb{P}_{f}$.
    To prove that our tripletGAN framework indeed works, there is one more thing we need to check. We have to ensure that $d_{\mathcal{F}}(\mathbb{P}, \mathbb{Q})=0$ implies $\mathbb{P}$ equals to $\mathbb{Q}$, which has not been proved in previous work like EM-distance and TV-distance does.
    \newtheorem{lemma}{Lemma}
    \begin{lemma}\label{lemma1}
        Suppose $S$ is a measurable set in $R^m$, and $m(S)>0$. Then $S$ can be represented as the union of two disjoint measurable sets with positive measure.
    \end{lemma}
    \emph{Proof:}\\
        \par{~~}
        We only consider the case when $m=1$, but the proof can be generalized without difficulty.
        Let $f:\mathbb{R}\mapsto \mathbb{R}^+$, $f(x) = m(S\cap [-x, x])$. It's easy to validate that $f$ is continuous, and $\lim_{x\to\infty}f(x)=m(S)$, $m(S)$ refers to the Lebseque measure of $S$. So for arbitrary $\epsilon$, $0<\epsilon<m(S)$, there exists a $C$ s.t. $m(S\cap [-C, C])=\epsilon$. So $S\cap [-C, C]$ and $S\cap (\mathbb{R}\setminus[-C, C])$ are the two sets we seek.
    \newtheorem{theorem}{Theorem}
    \begin{theorem}
        Suppose $\mathbb{P}$, $\mathbb{Q}$ are distributions over $\mathbb{R}^m$. $\mathcal{F}=\{f\mid f(x): \mathbb{R}^m\mapsto S^n , f\ is\ measurable\}$, $d_{\mathcal{F}}(\mathbb{P}, \mathbb{Q})=\sup\limits_{f\in \mathcal{F}}\{\mathop{\mathbb{E}}\limits_{\substack{y\sim\mathbb{P}\\ x\sim\mathbb{Q}}}\lVert f(x)-f(y)\rVert-\mathop{\mathbb{E}}\limits_{x_1, ~x_2\sim\mathbb{Q}}\lVert f(x)-f(y)\rVert\}$. Assume both $\mathbb{P}$ and $\mathbb{Q}$ have density function, denoted as $p(x)$ and $q(x)$ respectly, then $d_{\mathcal{F}}(\mathbb{P}, \mathbb{Q})=0$ if and only if $\mathbb{P}=\mathbb{Q}$.
    \end{theorem}
    \emph{Proof:}\\
    \par{~~}
    The necessity part of this theorem is obvious. We then focus to prove the opposite side.\\
    \[
        \begin{aligned}
        d_{\mathcal{F}}(\mathbb{P}, \mathbb{Q})=\sup\limits_{f\in \mathcal{F}}&\{\int_{\mathbb{R}\times \mathbb{R}}\lVert f(x)-f(y)\rVert p(x)q(y)dxdy-\\
        &\int_{\mathbb{R}\times \mathbb{R}}\lVert f(x)-f(y)\rVert p(x)p(y)dxdy\}\\
        =\sup\limits_{f\in \mathcal{F}}&\int_{\mathbb{R}\times \mathbb{R}}\lVert f(x)-f(y)\rVert p(x)(q(y)-p(y))dxdy
        \end{aligned}
    \]
    Since both $p(x)$ and $q(x)$ are measurable, so ${S_1}=\{q(y)-p(y)>0\}$ and ${S_2=\{p(x)>0\}}$ are measurable sets respectly. Suppose $\mathbb{P}\neq\mathbb{Q}$, then it amounts to find a $f_0\in \mathcal{F}$ s.t $\int_{\mathbb{R}\times \mathbb{R}}\lVert f(x)-f(y)\rVert p(x)(q(y)-p(y))dxdy>0$.\\
    Firstly, because
    \[
        \begin{aligned}
        \int_{\mathbb{R}^m}(q(x)-p(x))dx = &\int_{q(y)-p(y)>0}(q(x)-p(x))dx+\\&\int_{q(y)-p(y)<0}(q(x)-p(x))dx=0 ,
        \end{aligned}
    \]
    so $m(S_1)$ must greater than $0$, if otherwise, then both $m(S_1)$ and $m(\{q(y)-p(y)<0\})$ are $0$, contradict to the fact that $\mathbb{P}\neq\mathbb{Q}$. Moreover, $m(S_2)>0$ out of similiar reason.\\
    If $m(S_1\cap S_2)=0$, hence $S_2\subset \{p(y)=0\}$. Then we choose $f$ as follows:\\
    Let $z_0$ be a random point in $S^n$, let $f(x)=z_0$ for all $x$ in $S_1$ and $f(x)=-z_0$ for all x in $S_2$, and $f(x)=0$ for all other $x$ in $\mathbb{R}^m$.\\
    Then
    \[
        \begin{aligned}
            &\int_{\mathbb{R}\times \mathbb{R}}\lVert f(x)-f(y)\rVert p(x)(q(y)-p(y))dxdy\\
            =&\int_{S_1\times S_2}\lVert f(x)-f(y)\rVert p(x)(q(y)-p(y))dxdy\\
            &\int_{S_2\times S_1}\lVert f(x)-f(y)\rVert p(x)(q(y)-p(y))dxdy\\
            =&\int_{S_1\times S_2} 2 p(x)(q(y)-p(y))dxdy\\
            &\int_{S_2\times S_1} 2 p(x)(q(y)-p(y))dxdy>0\\
        \end{aligned}
    \]
    Thus $d_{\mathcal{F}}(\mathbb{P}, \mathbb{Q})>0$, contradict to previous hypothesis.\\
    If $m(S_1\cap S_2)>0$, according to lemma \ref{lemma1}, there exists $S_1^{\prime}$ and $S_2^{\prime}$ both with positive measure, and $S_1\cap S_2 = S_1^{\prime} \cup S_2^{\prime}$. Let $z_0$ be a random point in $S^n$, let $f(x)=z_0$ for all $x$ in $S_1^{\prime}$ and $f(x)=-z_0$ for all x in $S_2^{\prime}$, and $f(x)=0$ for all other $x$ in $\mathbb{R}^m$, then we will yield similiar result with previous case. The proof is completed.\\

    This theorem guaranteed when we minimized previous mentioned IPM between two joint distributions, we are driving $\mathbb{P}_f$ to become the same as $\mathbb{P}_r$, not any distribution else.
    \subsection{Training critic with hard examples}

    As pointed out in \cite{arjovsky2017towards} and \cite{arjovsky2017wasserstein}, the vanilla GAN as well as many other IPM based GAN suffer from gradients missing and degenerated discriminator problem. So it is the same with our proposed tripletGAN. 
    Recall the critic loss of tripletGAN is:
    \begin{equation}
     \begin{aligned}
        \mathscr{L}_c = \mathop{\mathbb{E}}\limits_{\substack{y\sim\mathbb{P}_{r},\\ z_1\sim\mathbb{N}}}\lVert f(y)-f(G(z_1))\rVert-\mathop{\mathbb{E}}\limits_{z_1, ~z_2\sim\mathbb{N}}\lVert f(G(z_1))-f(G(z_2))\rVert
        \end{aligned}
    \end{equation}
    Since the supports of $P_{real}$ and $P_{fake}$ are disjointed except for a zero-measure set, the critic $f$ are free to assign different values to the supports of two distribution. In other words, the critic $f$ can separate the samples of two distributions in base space accurately. Hence, to find the optimal $f$, we only need to find out the two values that $f$ assigns upon the two supports respectively. Because the value of $f$ lies on $S^n$, the former part of $\mathscr{L}_c$ will yield maximum if and only if $f$ maps samples from two distributions to two antipodal points of $S^n$. While the latter part would be maximized if and only if $f$ maps samples from $P_{fake}$ to a fixed point in $S^n$. The optimal $f$ is thus a degenerated function mapping samples to two fixed antipodal points.
    Since optimal $f$ maps all fake samples to a single point, of course, it gives no gradients to $G$, and that is the failure occasion what we want to prevent. So we proposed to only update $f$ using triplets which are hard for $f$ to separate. Exactly like what is stated in \ref{tripletloss}triplet loss section, say now we set the threshold to be $c$, then the clipped critic loss is:
    \[
        \small\begin{aligned}
            \mathscr{L}_t=\mathop{\mathbb{E}}_{\substack{y\sim\mathbb{P}_{r}\\ x_1,~x_2\sim\mathbb{P}_{f}}}\left[-\lVert f(y)-f(x_1)\rVert+\lVert f(x_1)-f(x_2)\rVert+c\right]_{+}\\
        \end{aligned}
    \]
    Intuitively, under clipped loss, the critic is encouraged to push the true embedding to be closer to fake embedding than the distance between fake embeddings by a margin of $c$. If the threshold $c$ is relatively large, say, exceeding $\pi$, then the optimal $f$ would still be the same as before for the loss of critic does not change at all. When $c$ is a little smaller than $\pi$, it's expected that optimal $f$ maps samples from two distribution to two small clusters in the poles of the sphere. the optimal $f$ is no longer degenerated rather map true and fake samples to two clusters in $S^n$. \\
    Because our framework organize samples as triplets, so when some triplets stop to pass gradient, there are often other triplets in the same batch responsible to update $f$. While in vanilla GAN, samples are treated equally as a minibatch, if some threshold is set on the GAN loss, then the discriminator would get virtually no gradients when loss exceeds the threshold.\\
    In our the following experiments, steady behaviors of both generator and critic loss are observed.\\
    \subsection{Triplet loss allows for diversity of fake samples}
    A critical difference between tripletGAN and vanilla GAN is, in vanilla GAN, discriminator only need to look at a single sample to decide which distribution this sample comes from, while in tripletGAN critic's task is to separate samples in a triplet away. A direct result out of this difference is, in vanilla GAN, it is enough for the generator to generate only one high probability sample to fool discriminator, there is no motivation for the generator to escape this situation. Meanwhile, in tripletGAN, the generator is able to compare two samples it emits, and are encouraged to generate them differently.
    Intuitively, the second item in the loss that generator is meant to minimize, $-\mathop{\mathbb{E}}\limits_{x,~y\sim\mathbb{Q}}\lVert f(x)-f(y)\rVert$, encourages embeddings of all samples in the same batch to differ with each other, so generator tends to explore more modes.\\
    Several experiments are conducted to prove this practically.
    \subsection{Algorithm}
    We present here the algorithm \ref{alg:framework} to train tripletGAN with $\mathscr{L}_t$. Note there are many ways to sample triplets from true and fake samples, suppose we have two batches $\{x_i\}$ and $\{G(z_i)\}$ with size $B$ now, each sampled from data and generated distribution separately, we construct triplets as $\left(G(z_i), G(z_j), x_i\right)$, $0<i,\,j<B,\, i\neq j$. So we have $B(B-1)$ triplets out of two batches. This sampling method is not as expensive as to give ten thousands of triplets, and still maintains the property to differ fake samples in the same batch. Other sampling methods are also generally feasible, as long as they produce a set of valid samples for $(x_1, x_2, y),\,y\sim\mathbb{P}_{r},\, x_1,~x_2\sim\mathbb{P}_{f}$. For ease of formalizing, we use vanilla SGD in the description of the algorithm, in real scenario it's easy to convert to other variants.
    \begin{algorithm}[t]
        \caption{Triplet GAN}\label{alg:framework}
        \begin{algorithmic}[1]
        \small
        \REQUIRE
        generator $G_{\theta_g}$; noise $z$; discriminator $f_{\theta_c}$; dataset $\mathcal{S}=\left\{X_{i}\right\}$, learning rate $\epsilon$, batch size $N$, threshold $c$

        \STATE
        Initialize $\theta_g$, $\theta_c$, $D_\phi$.
        \STATE

        \REPEAT
        \STATE
            Sample a minibatch from $\mathcal{S}$, yield $x_i$, $i=1...N$\\
        \STATE
            Sample a minibatch from $z$, yield $z_i$, $i=1...N$\\
        \STATE
            Sample triplets from these two minibatch, yield $\mathcal{T}=\{t_{i,j}\mid\{t_{i,j}=(G_{\theta_g}(z_i), G_{\theta_g}(z_j)), x_i\}$, $i=1...N, j=1...N$
        \STATE
            $\mathcal{L}_{c}(\theta_g,\theta_c)\gets\textstyle\dfrac{1}{N(N-1)}\sum\limits^N_{i=1}\sum\limits^N_{j=1}\min(\lVert f_{\theta_c}(x_i)-f_{\theta_c}(G_{\theta_g}(z_i))\rVert-\lVert f_{\theta_c}(G_{\theta_g}(z_i))-f_{\theta_c}(G_{\theta_g}(z_j))\rVert, c)$
        \STATE
            $\theta_c \gets \theta_c$+$\epsilon\nabla_{\theta_c}\mathcal{L}_{c}(\theta_g,\theta_c)$
        \STATE
            $\mathcal{L}_{g}(\theta_g,\theta_c)\gets\textstyle\dfrac{1}{N(N-1)}\sum\nolimits^N_{i=1}\sum\nolimits^N_{j=1}\lVert f_{\theta_c}(x_i)-f_{\theta_c}(G_{\theta_g}(z_i))\rVert-\lVert f_{\theta_c}(G_{\theta_g}(z_i))-f_{\theta_c}(G_{\theta_g}(z_j))\rVert$
        \STATE
            $\theta_g \gets \theta_g$-$\epsilon\nabla_{\theta_g}\mathcal{L}_{g}(\theta_g,\theta_c)$
        \UNTIL{Triplet GAN converges}
        \end{algorithmic}
    \end{algorithm}
    \begin{figure*}[htb]
        \vspace{-4pt}
        \centering
        \includegraphics[width=0.90\textwidth]{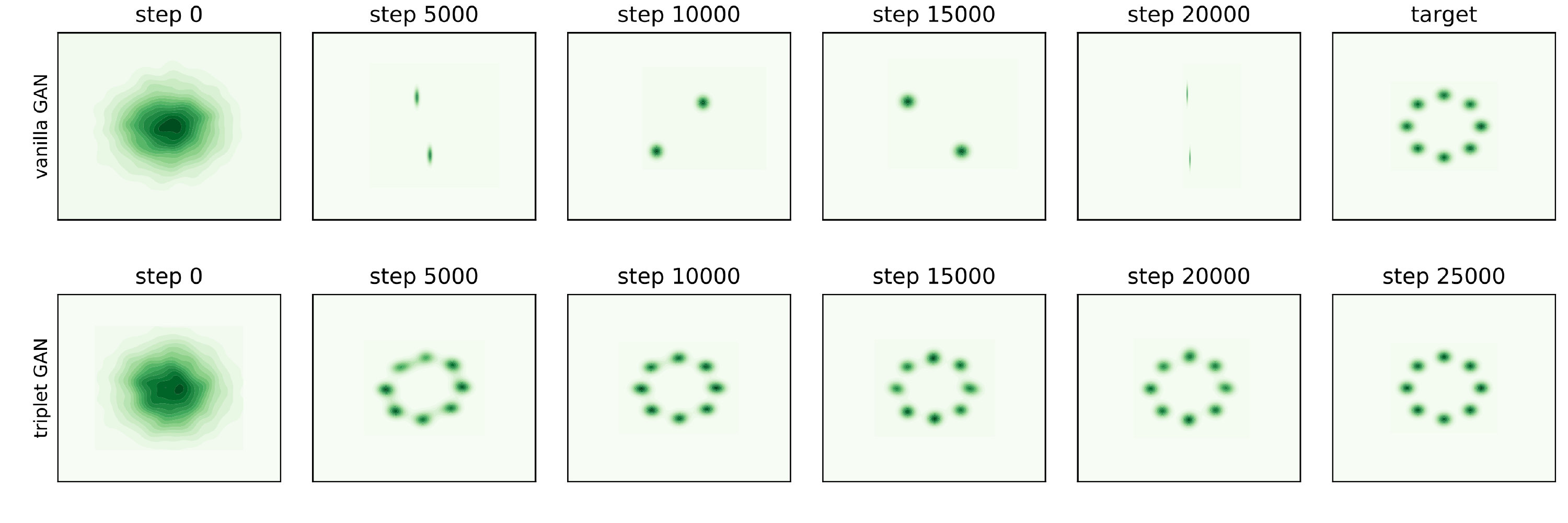}
        \vspace{-8pt}
        \caption{Mixed Gaussian experiment results, shown as heatmaps of generated distributions. The rightmost column shows the heatmap of target distribution, i.e. 8 mixed Gaussian distributions around a circle. The top row is the result of vanilla GAN and the second row is of tripletGAN. It's ready to see vanilla GAN are only able to capture few modes, and the means of each captured mode rotate as training goes on. While tripletGAN is able to nearly find all modes at the begining of training.}
        \label{fig:GAN_gaussian}
        \vspace{-10pt}
    \end{figure*}
    \begin{figure}[htb]
        \vspace{-10pt}
        \centering
        \subfigure{
            \begin{minipage}{7cm}
            \centering                
            \includegraphics[width=0.9\textwidth]{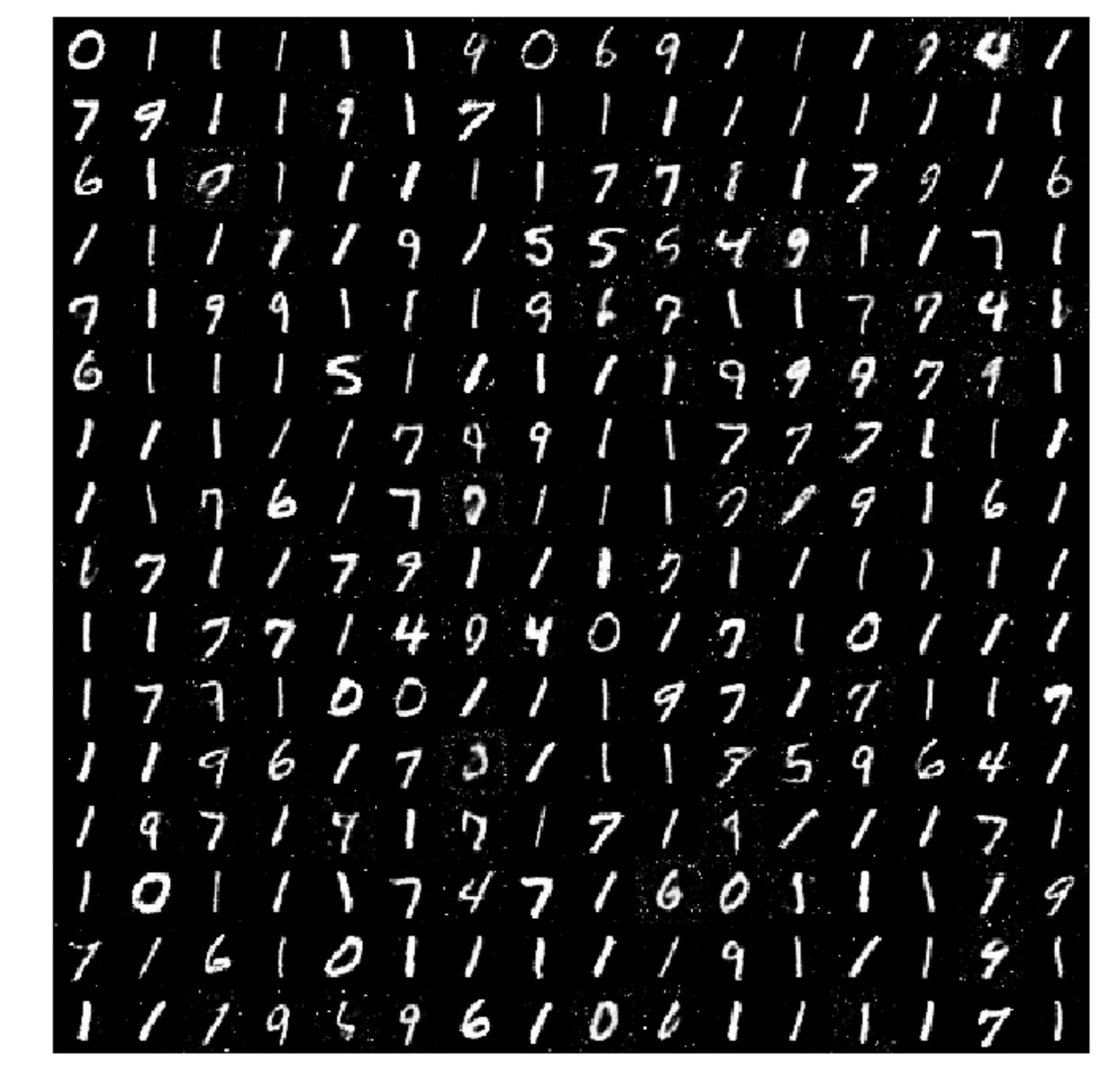}
            \caption{Results of GAN trained on MNIST. Most of samples are one-like digits.}
            \label{fig:GAN_mnist}
            \end{minipage}                
        }
        \vspace{-8pt}
        \vspace{-4pt}
        \centering
        \subfigure{
            \begin{minipage}{7cm}
            \centering            
            \includegraphics[width=0.9\textwidth]{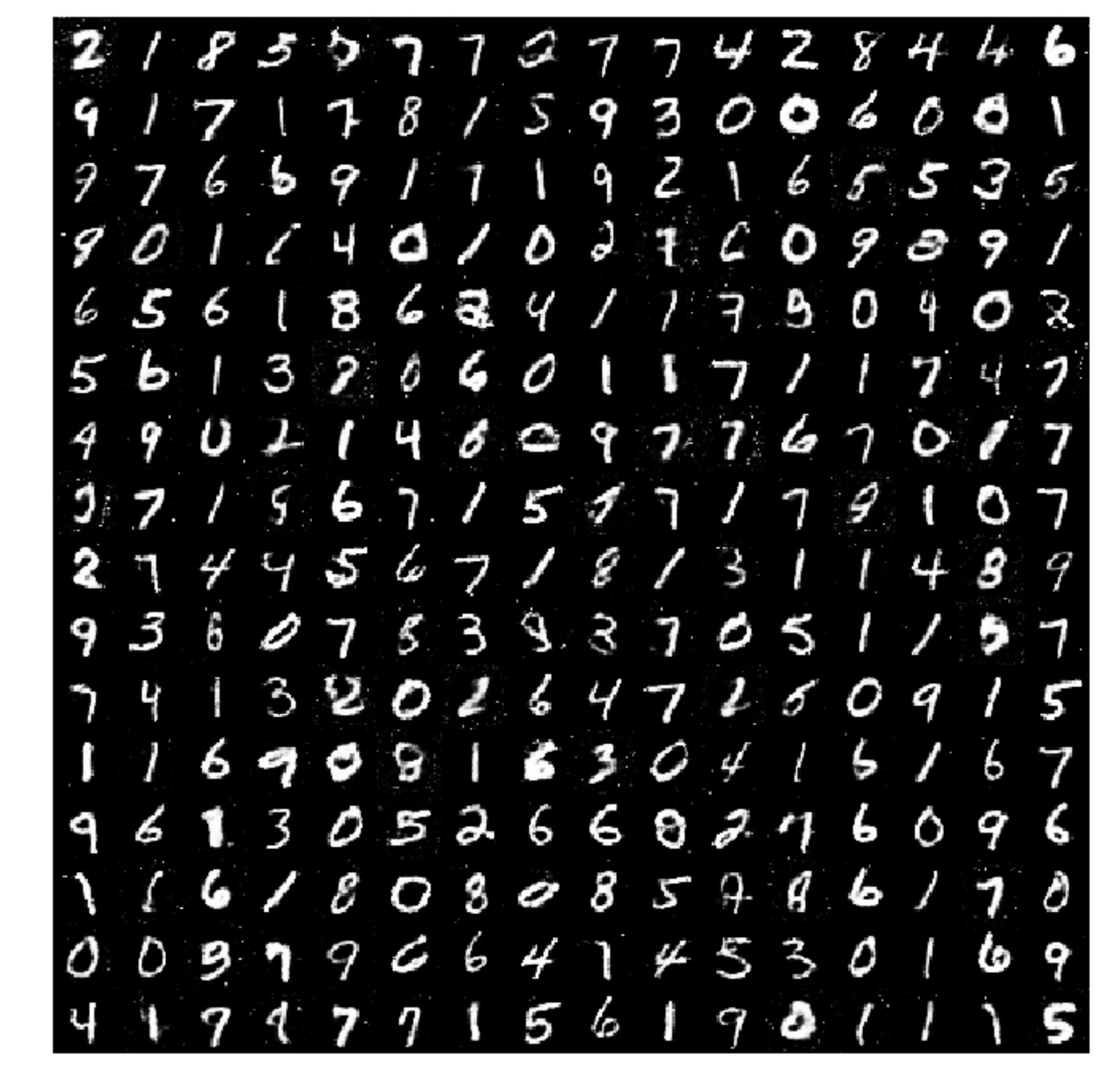}
            \caption{Results of triplet GAN trained on MNIST}
            \label{fig:triplet_GAN_mnist}
            \end{minipage}
        }
        \vspace{10pt}
    \end{figure}

    \section{Experiments}
    To prove the effectiveness of our proposed tripletGAN on preventing mode collapse against vanilla GAN, we train tripletGAN on various datasets, including a toy synthetic Gaussian distribution, MNIST digit dataset, and cropped CelebA \cite{liu2015faceattributes}. In all datasets, we observed a superior diversity of samples generated from tripletGAN over vanilla GAN.
    \subsection{Mixed Gaussian}
    Firstly, an experiment of mode recovery is performed, with a setting similar to previous works who seek to resolve mode collapse problem \cite{metz2016unrolled,che2016mode}. The target of tripletGAN is set to a mixed Gaussian distribution, where 8 same single mode Gaussian distributions with standard deviation as 0.01, are arranged as a circle of radius 1 in a 2D plane, and the task for it is to cover all modes. The same experiment in vanilla GAN with exactly the same setting is also performed in contrast. The final feature of $f$ in tripletGAN needs to be normalized by its L2 norm, to guarantee itself lying on $S^n$, $n=16$ here. And the threshold for critic loss is set to be 0.5. The implementation details are listed in Supplement Materials. Results are shown in Figure \ref{fig:GAN_gaussian}.\\
    Among all experiments, we use the length of the minor arc connecting two points in $S^n$ to act as the norm in $S^n$, but it actually equals to use unsquared L2 norm directly since they differ only upon a monotonic increasing function. All the arguments still hold regardless of the choice of norm.\\
    It can be seen from Figure \ref{fig:GAN_gaussian} that vanilla GAN can only locate two modes at the beginning of training, and generate samples oscillating around all modes afterward. Whereas in tripletGAN generated distribution are able to cover nearly all modes at an early stage of training.

    \begin{table}[!t]
        \label{tab:entropy}
        \centering
        \begin{tabular}{|l|l|l|l|}
            \hline
            & Vanilla-GAN & EBGAN & tripletGAN \\ \hline
            Entropy & 1.469       & 2.046 & 2.073      \\ \hline
            L2      & 0.502       & 0.060 & 0.057      \\ \hline
        \end{tabular}
        \caption{The entropys and L2 distances of the class distribution on MNIST. For entropy, the more close to $ln(10)=2.30$ the better. For L2 distance to uniform distribution, the less the better.}
    \end{table}

    \subsection{MNIST}
    MNIST is employed in our experiment to evaluate the performance of our model against other GANs on image generation task. Both the generator and the critic are simply feed-forward MLP rather than a convolutional network in order to test the mode coverage of each model in a simple architecture. We show the results in Figure \ref{fig:GAN_mnist} and Figure \ref{fig:triplet_GAN_mnist} respectively. It is obvious that vanilla GAN generates more of digit "1" and cares little about other digits, but triplet GAN can generate a rather balanced coverage of ten digits. A classifier is trained on MNIST to obtain the generated class distribution. The variety of generation results is evaluated in terms of entropy and L2 distance from a uniform distribution for vanilla GAN, EBGAN, and tripletGAN, as shown in \ref{tab:entropy}. TripletGAN shows richer variety than other models, especially than vanilla GAN.\\
    The reason for vanilla GAN tends to generate 1 are due to the overall simple architecture of both generator and discriminator, intuitively, vanilla GAN tends to do easy things in the current framework. But triplet GAN is able to generate hard samples such as 4 and 5, even though it shares the same network architecture with vanilla GAN.\\
    \begin{figure}[htb]
        \vspace{-10pt}
        \centering
        \subfigure{
            \begin{minipage}{7cm}
            \centering                
        \includegraphics[width=0.9\textwidth]{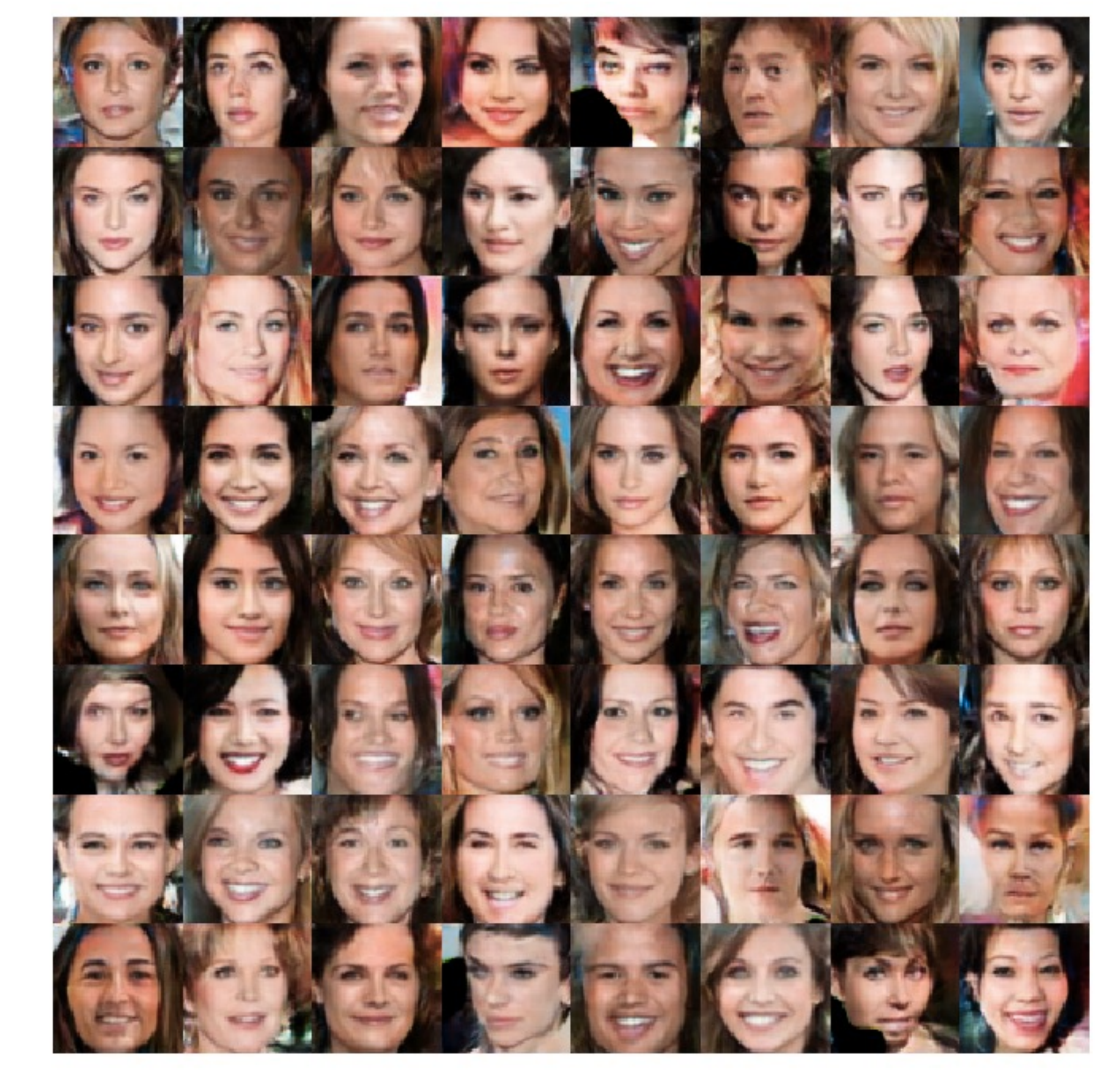}
        \caption{CelebA results for vanilla GAN. 7 male faces out of 64 samples.}
        \label{fig:GAN_celeba}
            \end{minipage}                
        }
        \vspace{-8pt}
        \vspace{-4pt}
        \centering
        \subfigure{
            \begin{minipage}{7cm}
            \centering            
        \includegraphics[width=0.9\textwidth]{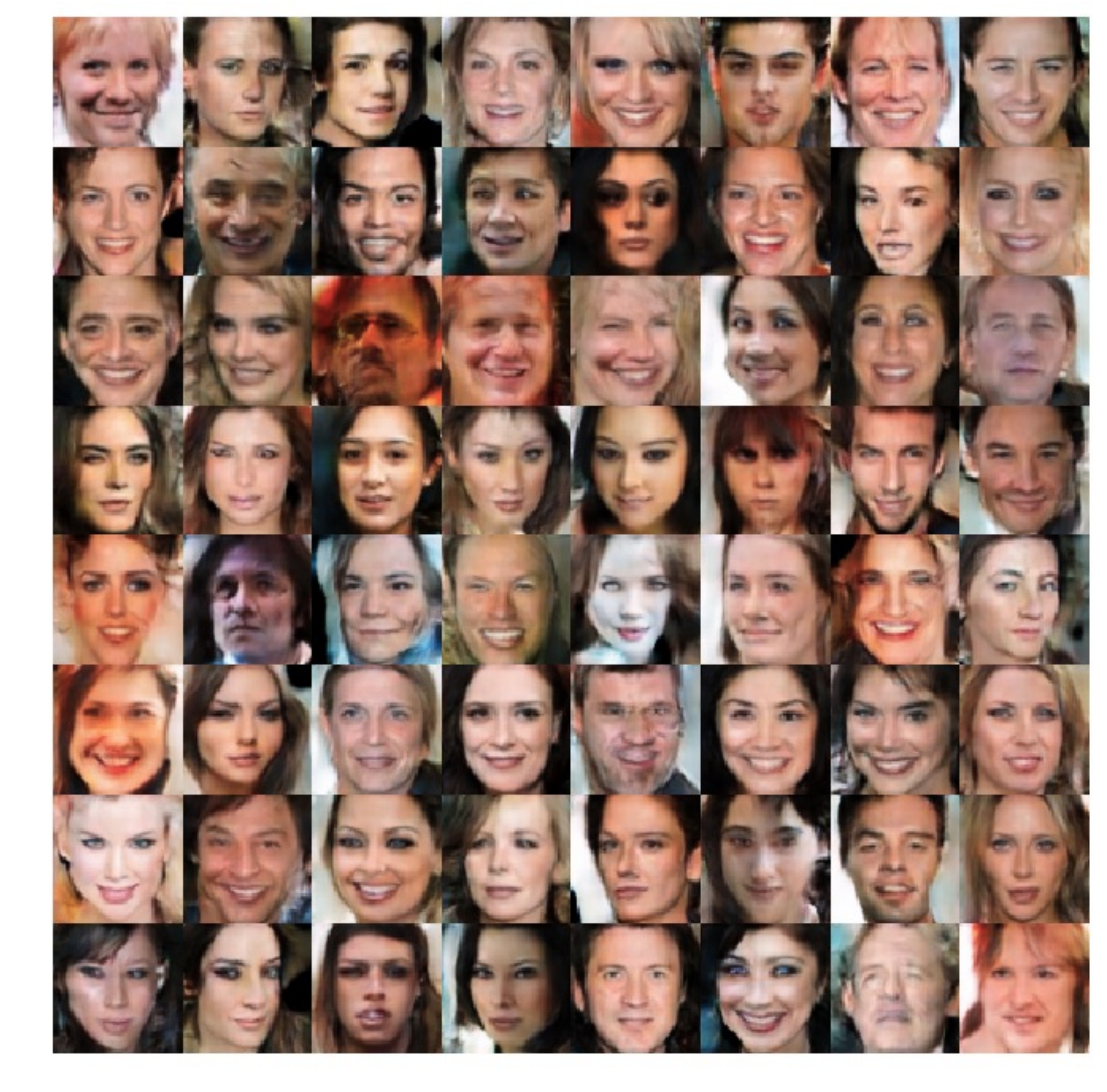}
        \caption{CelebA results for triplet GAN. 21 male faces out of 64 samples.}
        \label{fig:triGAN_celeba}
            \end{minipage}
        }
        \vspace{10pt}
    \end{figure}
    \begin{figure}[htb]
        \centering
        \includegraphics[width=0.9\textwidth]{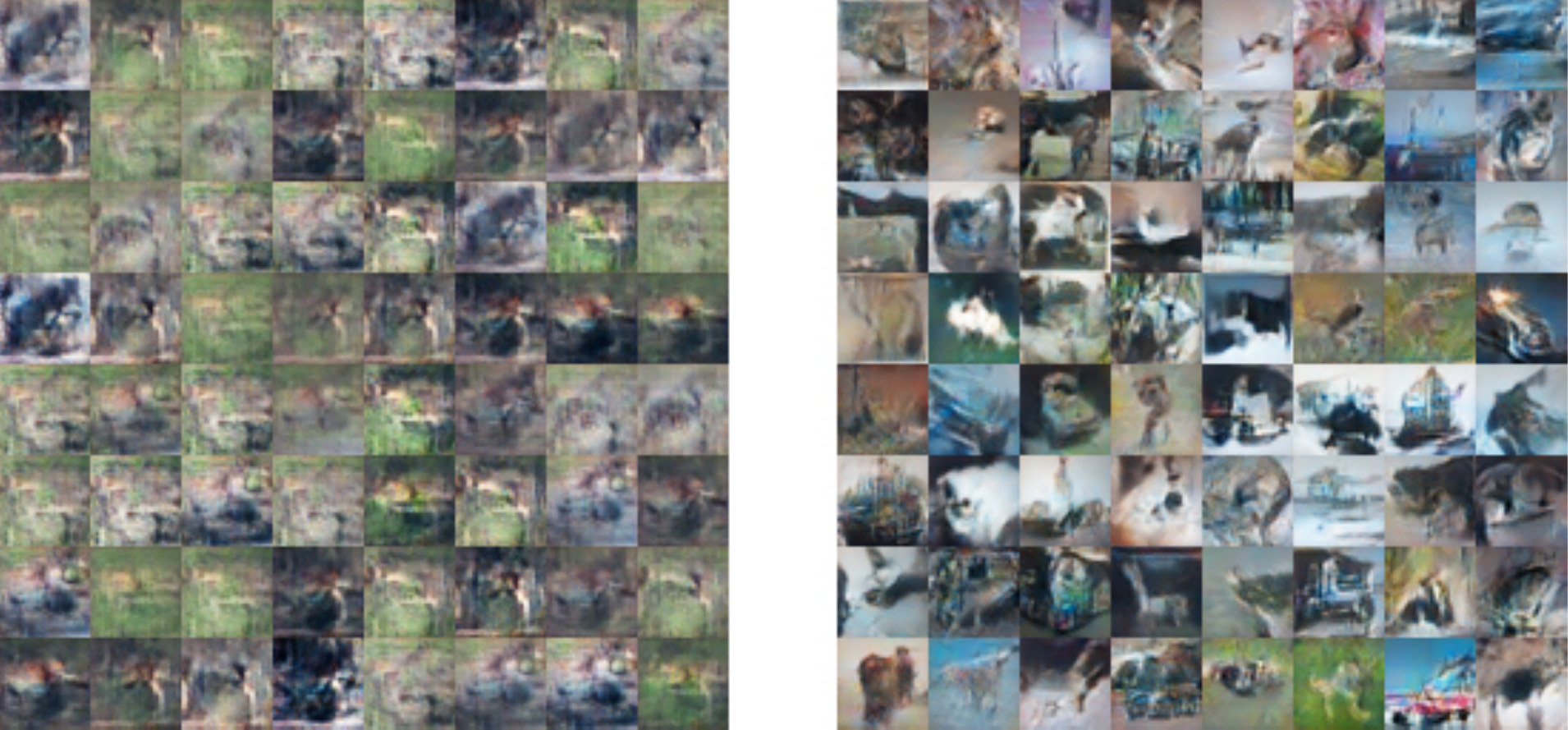}
        \caption{Comparison of generated samples on CIFAR-10 of vanilla GAN(left) and tripletGAN(right).}
        \label{fig:cifar_compare}
        \vspace{-10pt}
    \end{figure}
    \subsection{CIFAR-10 and STL-10}

    To address the sample quality of our model, we also test our model on CIFAR-10 \cite{krizhevsky2009learning} and STL-10 \cite{adam2011stl} using inception score as a criterion. The CIFAR-10 contains $60,000$ $32\times 32$ images from $10$ classes. STL-10 has $100,000$ unlabeled $96\times 96$ images and are resized to $64\times 64$ for training. The network structure is DCGAN-like, but no normalization trick such as batch normalization is performed to stabilize training. Samples from vanilla GAN and our model of CIFAR-10 are shown in Figure \ref{fig:cifar_compare}. Samples from vanilla GAN clearly stuck in a bad mode, indicating a dependence on normalization tricks, while our model is still able to generate decent images. We also compare our model with EBGAN and WGAN in inception score\cite{salimans2016improved}, the results are shown in \ref{tab:inception}.

    \begin{table*}[!t]
        \label{tab:inception}
        \centering
        \begin{tabular}{|l|l|l|l|l|}
            \hline
            inception score & Vanilla-GAN & WGAN       & EBGAN      & tripletGAN \\ \hline
            CIFAR-10        & 1.85+-0.04  & 2.33+-0.04 & 3.83+-0.13 & 4.42+-0.22 \\ \hline
            STL-10          & 5.21+-0.13  &            & 5.73+-0.20 & 6.44+-0.31 \\ \hline
        \end{tabular}
        \caption{The inception score from various models on CIFAR-10}
    \end{table*}

    \subsection{CelebA}
\begin{figure}[htb]
        \vspace{-4pt}
        \centering
        \includegraphics[width=0.45\textwidth]{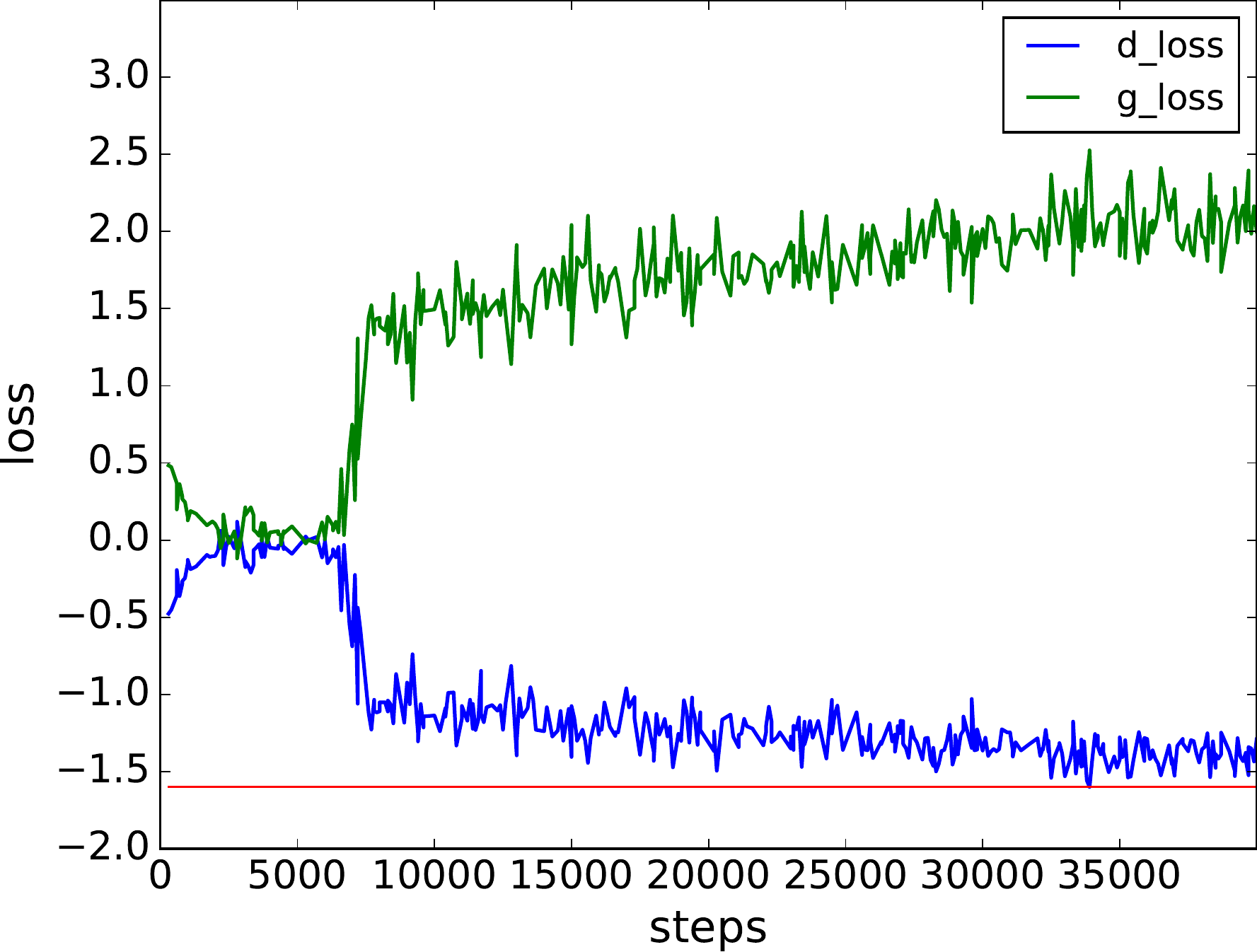}
        \vspace{-8pt}
        \caption{Loss curve of tripletGAN, where $c=1.6$ is marked in figure as the red line below.}
        \label{fig:loss}
        \vspace{-10pt}
    \end{figure}

    In order to test tripletGAN in a harder problem, such as human face generation, we train a vanilla GAN and tripletGAN on the CelebA dataset, where a face detection is performed and the face part is cropped to restrict our task on face generation. Our network architecture much resembles BEGAN's \cite{berthelot2017began}, which creates the state of the art in face generation currently. But we do not claim any superiority in image quality over other models, this experiment is only carried out to illustrate the sample diversity of our model.\\
    No batch normalization\cite{ioffe2015batch} is performed either in generator or discriminator's architecture, since batch normalization can be seen as a dirty way to force activations to differ with each other, which might hamper the fair comparison of the intrinsic property of model itself.\\
    The generated results of both models are shown in Figure \ref{fig:GAN_celeba} and Figure \ref{fig:triGAN_celeba}, these samples are all generated for the first time and not cherry-picked. Through observation we know, vanilla GAN, though overall generates pleasing images, is more inclined to produce women's faces, since they are often more smooth than men's faces, while the sex ratio of samples from our tripletGAN is more balanced.\\
    The loss curves of both generator and critic are plotted in \ref{fig:loss}. At the beginning, $G$ and $f$ are able to keep a relative balance between themselves, but afterwards, $f$ starts to degenerate, which is also the case in vanilla GAN. Whereas in tripletGAN, the d\_loss are then prevented to be saturated to $\pi$, the maximum of itself, meantime still is able to be updated through the rather hard triplets in current minibatch, allowing the training to continue even if the balance is broken.
    \section{Conclusion}
    In this paper, we propose a new training approach for GAN, called tripletGAN, by introducing triplet loss to adversarial learning. We show how it can be connected with Integral Probability Metric, and give a proof about the effectiveness of triplet loss in view of IPM. Furthermore, we argue that the form of triplet allows the generator to avoid mode collapse problem. To support this idea, we conduct several experiments in various datasets to illustrate that tripletGAN shows better mode coverage than vanilla GAN.\\
    Many possible improvements are to be studied thoroughly, since a lot of work has been done to improve the performance of triplet loss in face recognition and many other fields, such as adding soft margin, introducing another sample in triplet to form a quandruplet \cite{2017arXiv170401719C}, performing hard example mining to triplets set. Most of them are able to be transfered to our framework without much difficulty. It is also promising to build a conditional tripletGAN that does not convey labels to discriminator explicitly like original conditional GAN does, but rather using triplet loss with respect to different generated classes.
    \bibliographystyle{plain}
    \bibliography{mybib}
    \newpage
    \appendix
        \begin{appendices}

            \section{Connections with MMD and Wasserstein GAN}
            The MMD between $\mathbb{P}_r$ and $\mathbb{P}_f$ induced by a kernel $k$ can be written as:
            \[
                M_k(\mathbb{P}_r, \mathbb{P}_f)=\max_{k}\mathop{\mathbb{E}}_{x_1,x_2\sim\mathbb{P}_r}k(x_1, x_2)-2\mathop{\mathbb{E}}_{x\sim\mathbb{P}_r,~y\sim\mathbb{P}_f}k(x, y) + \mathop{\mathbb{E}}_{x_1,x_2\sim\mathbb{P}_f}k(x_1, x_2)
            \]
            Choose $k(x_1, x_2)$ to be $\lVert f(x_1)-f(x_2)\rVert$, where $\lVert\cdot\rVert$ is a norm on $S^n$. We will yield:
                \begin{align*}
                M_k(\mathbb{P}_r, \mathbb{P}_f)=\max_{k}&\mathop{\mathbb{E}}_{x_1,x_2\sim\mathbb{P}_r}k(x_1, x_2)-2\mathop{\mathbb{E}}_{x\sim\mathbb{P}_r,~y\sim\mathbb{P}_f}k(x, y) + \mathop{\mathbb{E}}_{x_1,x_2\sim\mathbb{P}_f}k(x_1, x_2)\\
                =\max_{f}&\mathop{\mathbb{E}}_{y_1,y_2\sim\mathbb{P}_r}\lVert f(y_1)-f(y_2)\rVert-2\mathop{\mathbb{E}}_{x\sim\mathbb{P}_r,~y\sim\mathbb{P}_f}\lVert f(x)-f(y)\rVert + \mathop{\mathbb{E}}_{x_1,x_2\sim\mathbb{P}_f}\lVert f(x_1)-f(x_2)\rVert\\
                =\max_{f}&(\mathop{\mathbb{E}}_{y_1,y_2\sim\mathbb{P}_r}\lVert f(y_1)-f(y_2)\rVert-\mathop{\mathbb{E}}_{x\sim\mathbb{P}_r,~y\sim\mathbb{P}_f}\lVert f(x)-f(y)\rVert) + \\&(\mathop{\mathbb{E}}_{x_1,x_2\sim\mathbb{P}_f}\lVert f(x_1)-f(x_2)\rVert-\mathop{\mathbb{E}}_{x\sim\mathbb{P}_r,~y\sim\mathbb{P}_f}\lVert f(x)-f(y)\rVert)
                \end{align*}
            It is ready to see the IPM induced by triplet loss is a simplified version of MMD with a particular kernel. The first term who regards to the encoding real samples is missed in triplet loss. Since this term is not used in generator update, whether it is added or not will not affect the context of our claim about sample variety.\\
            In that case, the difference between our work and \cite{2017arXiv170508584L} lies mainly in the kernel class chosen and our omission of the term regarding only to real samples. In short, if the position of fake and real class in triplet loss interchanged and added back to original loss, we will yield an exact form of MMD represented with an adversarial kernel. The fact that minimizing triplet loss is equal to find a kernel for MMD seems not documented in any publication.
            Note both MMD-GAN and Cramer GAN\cite{2017arXiv170510743B} introduce a gradient penalty on critics, while we simply use clipped loss but also produce decent results.

            On the other hand, if we choose $\phi(x)=\mathop{\mathbb{E}}_{x_1\sim\mathbb{P}_f}\lVert f(x_1)- f(x)\rVert$, then the objective of critic can then be written as:
            \[
                \mathscr{L}_c=\mathop{\mathbb{E}}\limits_{\substack{y\sim\mathbb{P}_{r}}}\phi(y)-\mathop{\mathbb{E}}\limits_{x\sim\mathbb{P}_{f}}\phi(x)
            \]
            If $\phi$ is chosen to be in a Lipshitz-1 function set, then the above loss will resemble the objective of Wasserstein GAN.
            \section{Experiments details}
                \subsection{Mixed Gaussian experiments details}
                Like what has been performed in \cite{metz2016unrolled,che2016mode}, we target our model to a mixed Gaussian distribution composed of 8 Gaussian distributions whose means are placed equally around a circle of radius 1 and standard deviation as 0.01.\\
                Note unless specified, all settings and hyper-parameters are shared between GAN and tripletGAN throughout all experiments. The latent vector that generator act upon is sampled from 128 dimension unit Gaussian distribution with every component independent with each other. The generator is composed of 3 fully-connected layers, each with hidden size of 128 and tanh as activation function, followed by a linear projection to 2 dimension. The critic consists of 3 same fully-connected layers with hidden size of 32 and activation function of tanh, plus a linear projection to a vector of feature size, where feature size equals to 1 if the model is vanilla GAN, 16 if it is tripletGAN. Note the final feature in critic of tripletGAN need to be normallized by its L2 norm to guarantee itself lying on $S^n$, $n=16$ here, and the threshold for critic loss is set to be 0.5.\\
                The training algorithm we adopt is Adam \cite{kingma2014adam}, with $\beta_1=0.5$ learning rate be 1e-3 and 2e-4 for generator and critic respectly. Batchsize is set to be 512. We train the whole framework 25000 steps in total.\\
                Among all experiments, we use the length of minor arc connnecting two points in $S^n$ to act as the norm in $S^n$, but it actually equals to use L2 norm directly since they differs only upon a monotonic increasing function.
                \subsection{MNIST experiments details}
                Our MNIST experiment share many settings with \cite{li2017distributional}. Latent vector is sampled from 128 dimension unit independent Gaussian distribution. The generator is a feed-forward fully-connected network consisting of 3 hidden layers of hidden size 256, 512, 1024, each with leaky relu activations, followed by a linear projection to 1024 dimension with tanh activation. Note the data in MNIST is of 28x28x1=784 dimension, but we pad 2 pixels of 0 around each image so it is of 32x32x1 dimension in our experiment. The critic, are pretty much the reverse of generator, having 3 hidden layers of hidden size 1024, 512, 256, followed by a linear projection to 1 or 16 dimension, depending on the model training on. The feature emitted by triplet's critic need to be normallized by L2 norm. The threshold of critic loss is 1.0.\\
                We use Adam as training algorithm, and set both learning rate as 5e-4, $\beta_1=0.5$. The batchsize is fixed to be 256 and we train the network for 100000 steps.
                \subsection{CelebA experiments details}
                We adopt an architecture similiar to \cite{berthelot2017began} for both vanilla GAN and tripletGAN training. The architecture details are listed in \ref{celeba-gen} and \ref{celeba-dis}. Note for tripletGAN critic maps samples to 16 dimension and place a l2-normalization on it. No batch normalization \cite{ioffe2015batch} or other normalization trick is applied. Most activation functions in both generator and critic are elu \cite{clevert2015fast}, which provide more smooth gradients than leaky relu. The threshold of critic loss is 1.6.\\
                We use adam algorithm and learning rate of 1e-4 to update all networks, use batch size of 64, and train for 40000 steps.

                \begin{table}[htb]
                \centering
                \caption{Generator architecture in CelebA experiment}
                \label{celeba-gen}
                \begin{tabular}{|c|c|c|c|c|c|}
                \hline
                                                    & channel of outputs & stride  & kernel size & activation function & output size   \\ \hline
                input $z\sim \mathbb{N}(0, I_{128})$ & \multicolumn{4}{c|}{}                                            & (128,)        \\ \hline
                Fully connected                      & 4x4x512            & \multicolumn{2}{c|}{} & elu                 & (4x4x512,)    \\ \hline
                reshape to (4, 4, 512)               & 512                & \multicolumn{3}{c|}{}                       & (4, 4, 512)   \\ \hline
                Bilinear resize                      & 512                & \multicolumn{3}{c|}{}                       & (8, 8, 512)   \\ \hline
                Convolution                          & 256                & 1       & 3           & elu                 & (8, 8, 256)   \\ \hline
                Bilinear resize                      & 256                & \multicolumn{3}{c|}{}                       & (16, 16, 256) \\ \hline
                Convolution                          & 128                & 1       & 3           & elu                 & (16, 16, 128) \\ \hline
                Bilinear resize                      & 128                & \multicolumn{3}{c|}{}                       & (32, 32, 128) \\ \hline
                Convolution                          & 64                 & 1       & 3           & elu                 & (32, 32, 64)  \\ \hline
                Bilinear resize                      & 64                 & \multicolumn{3}{c|}{}                       & (64, 64, 64)  \\ \hline
                Convolution                          & 32                 & 1       & 3           & elu                 & (64, 64, 32)  \\ \hline
                Convolution                          & 3                  & 1       & 3           & tanh                & (64, 64, 3)   \\ \hline
                \end{tabular}
                \end{table}
                \vbox{}
                \vbox{}
                \vbox{}    
                \begin{table}[htb]
                    \centering
                    \caption{Discriminator architecture in CelebA experiment}
                    \label{celeba-dis}
                    \begin{tabular}{|c|c|c|c|c|c|}
                    \hline
                                                        & channel of outputs & stride  & kernel size & activation function & output size   \\ \hline
                    Convolution                          & 64                 & 1       & 3           & elu                 & (64, 64, 64)  \\ \hline
                    Convolution                          & 64                 & 2       & 3           & elu                 & (32, 32, 64)  \\ \hline
                    Convolution                          & 192                & 2       & 3           & elu                 & (16, 16, 192)   \\ \hline
                    Convolution                          & 192                & 1       & 3           & elu                 & (16, 16, 192) \\ \hline
                    Convolution                          & 256                & 2       & 3           & elu                 & (8, 8, 256)  \\ \hline
                    Convolution                          & 256                & 1       & 3           & elu                 & (8, 8, 256)  \\ \hline
                    Convolution                          & 320                & 2       & 3           & elu                 & (4, 4, 320)  \\ \hline
                    Convolution                          & 320                & 1       & 3           & elu                 & (4, 4, 320)  \\ \hline
                    reshape to (5120,)                   & \multicolumn{4}{c|}{}                                            & (5120,)   \\ \hline
                    Fully connected                      & 1 or 16            & \multicolumn{2}{c|}{} & none                & (1,) or (16,)    \\ \hline
                    Normallize(optional)                 & 16                 & \multicolumn{3}{c|}{}                       &(16,)          \\ \hline
                    \end{tabular}
                    \end{table}
                \subsection{CIFAR-10}
                The network architecture we use in CIFAR-10 experiments is like DCGAN except we do not perform any normalization trick. The decoder of critic for EBGAN is the mirror of encoder, with convolution substituted by transposed convolution. We use adam algorithm and learning rate of 2e-4 to update all networks, use batch size of 256, and train for 100000 steps. The threshold of critic loss is 1.6 and encoding dim is 64.
                \subsection{STL-10}
                The network architectures are shown below. The threshold of critic loss is 1.6. We use adam algorithm and learning rate of 2e-4 for generator and 1e-4 for critic, batch size of 64, and train for 40000 steps.
                \begin{table}[tbp]
                \centering
                \caption{Generator architecture in STL-10 experiment}
                \label{stl-generator}
                \begin{tabular}{|c|c|c|c|c|c|}
                \hline
                                                     & channel of outputs & stride  & kernel size & activation function & output size   \\ \hline
                input $z\sim \mathbb{N}(0, I_{128})$ & \multicolumn{4}{c|}{}                                            & (128,)        \\ \hline
                Fully connected                      & 4x4x512            & \multicolumn{2}{c|}{} & lrelu               & (4x4x512,)    \\ \hline
                reshape to (4, 4, 512)               & 512                & \multicolumn{3}{c|}{}                       & (4, 4, 512)   \\ \hline
                Bilinear resize                      & 512                & \multicolumn{3}{c|}{}                       & (8, 8, 512)   \\ \hline
                Convolution                          & 256                & 1       & 3           & lrelu               & (8, 8, 256)   \\ \hline
                Bilinear resize                      & 256                & \multicolumn{3}{c|}{}                       & (16, 16, 256) \\ \hline
                Convolution                          & 128                & 1       & 3           & lrelu               & (16, 16, 128) \\ \hline
                Bilinear resize                      & 128                & \multicolumn{3}{c|}{}                       & (32, 32, 128) \\ \hline
                Convolution                          & 64                 & 1       & 3           & lrelu               & (32, 32, 64)  \\ \hline
                Bilinear resize                      & 64                 & \multicolumn{3}{c|}{}                       & (64, 64, 64)  \\ \hline
                Convolution                          & 32                 & 1       & 3           & lrelu               & (64, 64, 32)  \\ \hline
                Convolution                          & 3                  & 1       & 3           & tanh                & (64, 64, 3)   \\ \hline
                \end{tabular}
                \end{table}

                \begin{table}[tbp]
                \centering
                \caption{Critic architecture in STL-10 experiment}
                \label{stl-critic}
                \begin{tabular}{|c|c|c|c|c|c|}
                \hline
                                     & channel of outputs & stride  & kernel size & activation function & output size    \\ \hline
                Convolution          & 64                 & 1       & 3           & lrelu               & (64, 64, 64)   \\ \hline
                Convolution          & 64                 & 2       & 3           & lrelu               & (32, 32, 64)   \\ \hline
                Convolution          & 128                & 1       & 3           & lrelu               & (32, 32, 128)  \\ \hline
                Convolution          & 128                & 2       & 3           & lrelu               & (16, 16, 128)  \\ \hline
                Convolution          & 256                & 1       & 3           & lrelu               & (16, 16, 256)  \\ \hline
                Convolution          & 256                & 2       & 3           & lrelu               & (8, 8, 256)    \\ \hline
                Convolution          & 512                & 1       & 3           & lrelu               & (8, 8, 512)    \\ \hline
                Convolution          & 512                & 2       & 3           & lrelu               & (4, 4, 512)    \\ \hline
                reshape to (5120,)   & \multicolumn{4}{c|}{}                                            & (8192,)        \\ \hline
                Fully connected      & 1 or 128           & \multicolumn{2}{c|}{} & none                & (1,) or (128,) \\ \hline
                Normallize(optional) & 128                & \multicolumn{3}{c|}{}                       & (128,)         \\ \hline
                \end{tabular}
                \end{table}

        \end{appendices}

\end{document}